# Final Report on MITRE Evaluations for the DARPA Big Mechanism Program

Matthew Peterson
Tonia Korves
Christopher Garay
Robyn Kozierok
Lynette Hirschman

MITRE

This page intentionally left blank.

# Table of Contents

# Executive Summary

This report documents MITRE's role in the DARPA Big Mechanism program. Specifically, MITRE was tasked with designing evaluations for the program and running the evaluations. The program chose as its use case the mechanisms of cancer cell signaling. The program evolved through three phases:

1. Reading: acquiring knowledge of cancer cell signaling mechanisms by reading the scientific literature (at scale), integrating structured knowledge from curated databases and leveraging biologists' expert knowledge;

2. Assembly: assembling interactions into mechanisms to support the creation of executable models of these mechanisms;

3. Reasoning: reasoning over the mechanisms and running executable models to provide explanations for empirically observed phenomena, for example, predicting how particular cell lines or pathways would respond to specific drugs.

For each phase, MITRE's goal was to design an evaluation that would enable developers and the DARPA program managers to assess how well the systems were performing in the context of the chosen Big Mechanism task. MITRE adopted a phased approach, evaluating new capabilities with each new phase. The evaluations were designed to build incrementally, challenging the performer teams to add new capabilities to address the new requirements of the phase.

By the end of the program, systems were built to read at scale over the open access collection from PubMed Central and abstracts in PubMed.  Furthermore, semi-automated Big Mechanism systems were able to combine machine-reading at scale with expert knowledge to create mechanistic models of cell signaling networks.  These models were able to provide plausible explanations for experimental observations on the effect of specific drugs on cell lines: given twenty drug-treatment experimental results from a published paper[1],  the top two systems provided plausible explanations for seventeen and nineteen of these results in the Phase III evaluation.

Overall, the evaluations presented a number of challenges:

- Evaluating at a "whole-of-literature/whole-of-knowledge" scale

- Evaluating systems' ability to extract knowledge in an open world domain (biology) and in the absence of a "gold standard"

- Designing evaluations to take into account systems' multiple sources of input, including expert humans, expert-curated databases, and information automatically extracted from the literature

- Designing evaluation criteria for a multi-stage multi-technology complex process

This report describes the evaluation approach for the Big Mechanism program; it includes an appendix that describes the data sets, scoring criteria, and results for the phases of evaluation.

# Abstract

This report presents the evaluation approach developed for the DARPA Big Mechanism program, which aimed at developing computer systems that will read research papers, integrate the information into a computer model of cancer mechanisms, and frame new hypotheses. We employed an iterative, incremental approach to the evaluation of the program's three phases. In Phase I, we evaluated system and human teams' ability to "read with a model" to capture mechanistic information from the biomedical literature, integrated with information from expert curated biological databases. In Phase II we evaluated systems' ability to assemble fragments of information into a mechanistic model. The Phase III evaluation focused on systems' ability to provide explanations of experimental observations based on models assembled (largely automatically) by the "Big Mechanism" process. The evaluation for each phase built on earlier evaluations and guided developers towards creating capabilities for the new phase. The report describes our approach, including innovations such as a "reference set" (a curated data set limited to major findings of each paper) to assess the accuracy of systems in extracting mechanistic findings in the absence of a gold standard, and a method to evaluate model-based explanations of experimental data. Results of the evaluation and supporting materials are included in the appendices.

This page intentionally left blank.

# Introduction

The Defense Advanced Research Projects Agency (DARPA) created the Big Mechanism program to develop technology for constructing, understanding and reasoning about big, complicated systems [2], including automated reading (natural language processing), knowledge representation, and the ability to reason over explicit representations of mechanisms. Big Mechanism's focus on mechanistic reasoning contrasts with many of the current highest-performing AI techniques (such as neural networks as well as statistical methods like support vector machines and random forests) that are largely opaque to humans [3]. Ultimately, the adoption and effectiveness of AI tools will be limited by a system's ability (or inability) to explain its decisions, recommendations, and actions to its human partners. Particularly in cases where lives may be at stake, AI algorithms need to not only be able to make good recommendations, but to back them up with explanations that are understood by the human decision makers. The Big Mechanism focus on reasoning over mechanistic explanations could bring human-machine partnerships to bear on critical applications, such as designing new therapies for cancer or supporting tumor boards in making critical treatment decisions.

The Big Mechanism program has been structured in distinct stages. The first stage is to obtain information about causal relationships (network "fragments") from curated databases and relevant scientific literature. Because researchers cannot keep abreast of the huge volume of literature, automated reading for information on these fragments is critical. The second step is the assembly of these fragments into a network by machines. The third step is the parameterization of these networks to produce models that can run simulations and generate explanations. In turn, these models can be used to generate hypotheses that feed back into this process and can be experimentally tested.

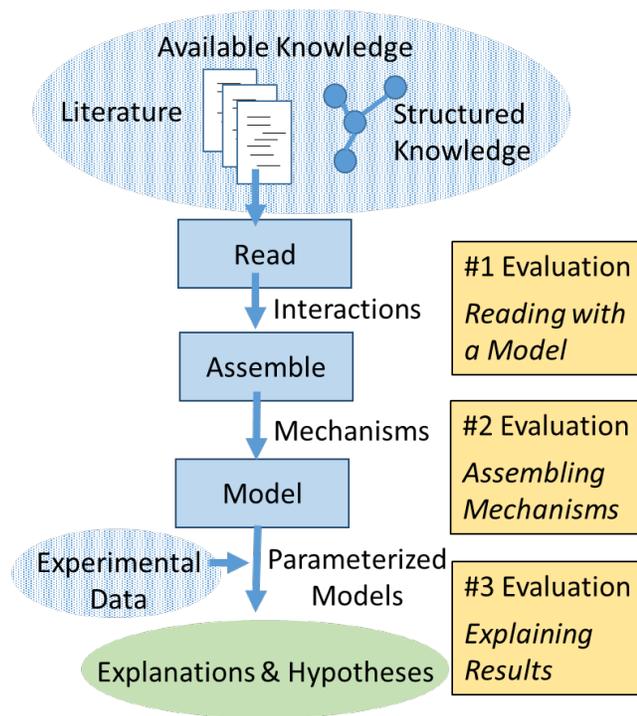

Figure 1 The Big Mechanism process and evaluations

The use case chosen for the Big Mechanism program was mechanistic modeling of cancer cell signaling, which can be used to explain the effects of drugs on cancer cells. This use case had several advantages to support the development of Big Mechanism systems, including the availability of structured vocabularies for relations and entities involved in these pathways; richly curated databases of signaling pathways; and over 1 million open access articles for machine reading. Cancer signaling was also a good use case because it has very complex mechanisms. Cancer signaling pathways regulate complex biological processes including cell growth, division, and migration; these pathways consist of proteins that interact to propagate and integrate environmental signals. Signals are transduced by adding and removing chemical groups to proteins, leading to activation or inhibition of protein activity.

Evaluation has been an integral part of the Big Mechanism program. The evaluation approach evolved organically to play the dual roles of focusing the research and providing assessment of the underlying technologies. There were three evaluations corresponding to the main steps in the Big Mechanism process (Figure 1). The remaining sections of the paper provide background on evaluation approaches; the specifics of the evaluations, highlighting the results and the challenges in designing evaluations to encompass the major goals of the Big Mechanism program; the lessons learned from this approach to evaluation; and a conclusion. The Appendices contain details about the evaluations for each phase, along with the data sets used in the evaluation, the evaluation criteria, and anonymized results.

## Evaluation Approach

A major challenge for the evaluation in the Big Mechanism program was to design appropriate evaluations that would both enable assessment of system progress in terms of the Big Mechanism challenge and guide research to meet the next phase of the challenge.

Current evaluation approaches, particularly for tasks involving human cognition, are usually framed as comparing an automated system's performance to that of expert humans. The most straightforward way to measure this is to define a representative task, ask humans to do it, collect the data, adjudicate the results to define a "gold standard" and then compare automated system performance to the human "gold standard". This has been the dominant evaluation paradigm, leading to the creation of rich annotated data sets and numerous challenge evaluations (open competitions) in language understanding. Evaluations include the early MUC evaluations from the '80s [4], the TREC (Text Retrieval) and TAC (Text Analysis) conferences, and a number of evaluations for natural language processing applied to biology [5]: BioCreative, BioNLP, and BioASQ. In addition, there is a long tradition of computational biology critical assessments, starting with the CASP protein structure competitions and more recently, the DREAM[6] and CAMDA evaluations, among others. This paradigm also underlies other competitions such as the Netflix Prize[7]. The Watson [8] success in Jeopardy was based on use of correct (gold standard) question/answer sets to train a system that could provide answers and learn a good strategy for responding.

Evaluation of progress on a complex task, such as that defined in Big Mechanisms, poses multiple challenges. For one, it may not be practical to achieve an expert consensus for reasons of cost or access to expertise or lack of consensus among experts – or because of incomplete knowledge in an "open world" scientific domain, where new knowledge is constantly being discovered. This was the challenge faced in the Big Mechanism program, but this is equally problematic for other complex, multi-scale applications. Second, the Big Mechanism problem involved working at scale, using all available resources. These resources could include carefully curated databases constructed by human experts, information automatically extracted from the published literature, and other kinds of expert input solicited in various ways, such as carefully constructed domain ontologies. This combination of factors has required the design of new evaluation approaches that go beyond the traditional gold-standard based approaches typically used for evaluation in natural language processing and machine learning.

One of the experiments in the Phase I evaluation used human curation to generate a "silver standard" set. This experiment had researchers manually extract interactions from text; the results showed great variability and poor interannotator agreement. This was due in part to differences in interpretation of the extraction task, as well as the difficulties that humans had in doing certain tasks, such as linking proteins to the correct identifiers in standard resources (for example, UniProt). It would have required extensive training of curators and significant time to produce a dataset of sufficient quality to use, even as a "silver standard". Therefore, we used MITRE expert biologists (co-authors MP, TK, CG) to judge performer output according to clearly specified criteria (documented in Appendix A), and where necessary, to generate small curated data sets.

## Evaluation Designs by Phase

The phase-specific evaluation approaches are outlined below:

- Phase I (focus on reading): Use of text-based evidence, where systems were required to "show their work" to enable semi-automated evaluation by domain experts. This allowed evaluators to quickly compare extracted information against the evidence provided as text excerpts.

- Phase II (focus on assembly of interactions into mechanisms): Creation of a carefully curated "reference set" of interactions, curated to represent major mechanistic findings from a small set of papers (used in Phase II). This allowed us to calculate basic measures such as reference set overlap (a proxy for recall) and precision for the systems.

- Phase III (focus on executing parameterized models and explaining the results):
    - Comparison between modeling output and experimental data in a manner similar to community challenges such as the DREAM challenges [6], based on using specific experimental conditions to parameterize models
    - Explanation evaluation: Because concordance between model predictions and experimental results did not specifically assess the explanatory power of mechanistic models, we developed an evaluation which focused on explanation, rather than prediction, based on system output that consisted of sets of explanatory interactions backed by provenance and evidence information.

To develop these evaluations, the MITRE team worked in close coordination with the DARPA program manager and the performer teams. The goal of the evaluations was to enable the program to make quantifiable progress on its objectives, with an emphasis on collaboration and sharing among performer teams. Performers were encouraged to form consortia at the kick-off meeting, and some of those consortia lasted for the duration of the program.

The evaluations were designed to be both incremental and iterative, with opportunities for performer feedback on design of evaluation tasks and submission standards. Each phase included multiple dry runs that enabled the performer teams to incrementally improve their systems and to ensure that the systems conformed to data exchange standards for evaluation. This also allowed the MITRE team to refine the evaluation criteria and the scoring methods. Performers were also encouraged to incorporate human expertise into the system where they thought it appropriate.

The approach applied in Big Mechanism proved successful on several fronts: it provided diagnostics to the performer teams; it encouraged sharing and collaboration, both among performers and between performer teams and MITRE. Each new level of evaluation focused performers' attention on the next step rather than on perfecting the preceding step, with an emphasis on applying additional knowledge sources and reasoning. This allowed performers in the program to make significant progress and to quickly leverage existing resources (expert knowledge, thousands of articles, curated databases) to build complex models that could explain empirical observations.

## Phase I: Reading with a Model

In Phase I, teams were challenged to develop systems that could rapidly read scientific papers for experimental results related to a model of cell signaling (Figure 2A). Systems were given full text primary research papers and asked to retrieve interactions between pairs of entities, with a focus on the types of entities and molecular mechanisms that are important in cancer signaling models (Figure 2B). For each interaction, systems had to provide text excerpts from the paper from which the interaction was extracted, to facilitate evaluation and to capture provenance. Because identifying entities consistently is important in assembling information, systems were required to assign standard database identifiers to entities. Systems were also given a cancer signaling model and asked to return only interactions that related to that model, either by corroborating, contradicting, extending, or adding specificity to an interaction already existing in the model. To encourage rapid, high-throughput system reading, systems were given just one week to process 1000 papers and had to return standardized output representing each interaction in a JSON format, dubbed an "index card" (Figure 2C). A summary of the index card content can be found in Appendix A. Finally, to encourage teams to explore human contributions in big mechanism systems, there was also a human-system condition in which system-generated index cards from 100 papers were filtered and/or edited by team scientists.

Index cards for ten papers per submission were reviewed manually to assess precision and throughput. For precision, interactions on index cards were considered correct if both participants and the interaction type were consistent with the evidence text on that index card, and the evidence text was about experimental results of the paper. Because the focus was on experimental results, interactions on index cards with evidence text based on background and methods were not evaluated. In addition, when there were index cards with the same interaction in a submission, only one was evaluated. Precision was then calculated as the number of correct interactions divided by the number of evaluated interactions. Throughput was estimated by dividing the number of correct cards by the number submitted for the papers reviewed, and then dividing this by the time given to generate the cards. Time for generation was seven days for system-only and three days for human-system.

Four teams built systems that processed all 1000 papers in a week and returned index cards for 71% to 90% of the papers. These systems generated 1.5 to 6.7 index cards per paper that correctly identified the participants and interaction type, with precision ranging from 0.23 to 0.63 (Figure 2D). Among index cards that correctly identified participants as text strings, the percent that had both participants correctly

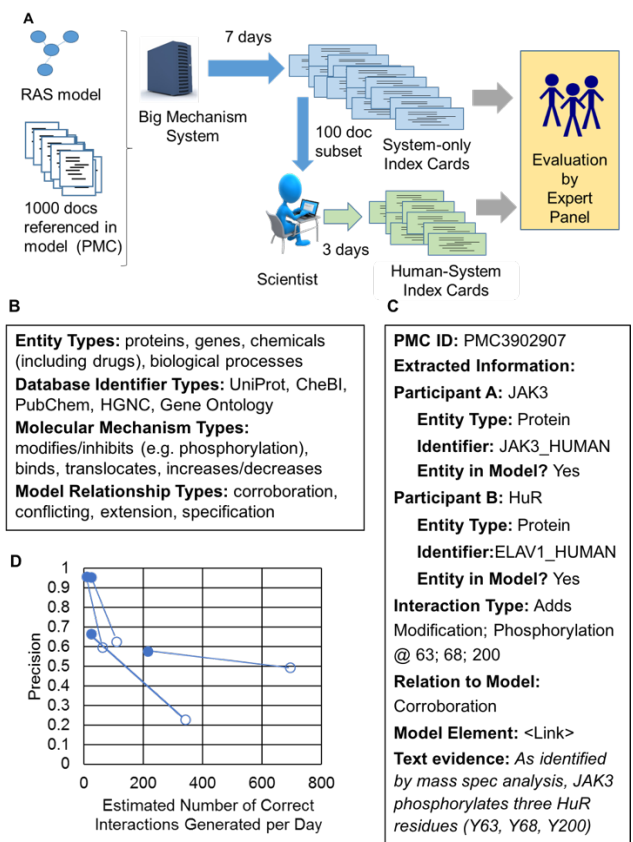

*Figure 2 (A) Phase I evaluation design. Teams were given PubMed Central open access papers and a cancer signaling model from Pathway Commons that included 2000 interactions within two hops of the RAS proteins. Systems extracted interactions related to the model and returned them in "index card" format. In the human-system condition, scientists filtered and/or edited system-generated index cards for 100 papers. B) Types of entities, database identifiers, molecular mechanisms, and model relationships that machines were asked to return. C) Example index card showing an interaction triple, represented as two molecules (participants A and B) and interaction type ("adds modification"). D) Precision versus number of correct interactions generated per day of effort. Open circles, machines only. Filled-in*

grounded to database identifiers ranged from 70 to 81%. When system-generated cards were revised by humans, precision improved at the expense of throughput (Figure 2D). All systems returned interactions that were related to the model, and three systems distinguished between corroboration and extension relationships.

To examine how the machine capabilities compared to those of humans, eight scientists with limited curation experience and training collectively curated 37 of the papers that machines read (see Appendix B for details). Scientists generated index cards that had higher precision than those of machines, with an average precision of 0.85 across scientists. However, on the scientists' index cards, grounding to database identifiers was much less accurate than for machines; of the cards that had correct participants, just 31% had correct database identifiers for both entities. When the scientists' index cards were run through the machine systems to reassign identifiers to entities, 73% of the index cards generated were correct for participants, interaction types, and database identifiers. These experiments suggested that a combination of current strengths of machines and humans, either via machine editing of human-generated cards or human filtering of machine-generated cards as in Figure 2D, could improve curation accuracy.

While these results demonstrated that the systems could read at scale for information related to a model, the evaluation also revealed shortcomings in the machine output for model building. First, precision for interactions was not high. Second, accuracy for database identifiers assigned to entities, though better than that from scientist curators, still included many errors that could be a problem for integrating information for model building. Third, machines generated many excess index cards; less than half of the index cards submitted by each team had unique interactions about experimental results from the papers. Many of the interactions returned were based on background information, methods, or minor points from a study, rather than major experimental results that are desirable for model building. In addition, system output contained many duplicate and identical interactions; consolidation of this information would be necessary for model building. Consequently, the Phase II challenge was designed to improve precision, prioritization, and assembly of information.

## Phase II: Paper-Level Assembly

In the Phase II challenge, systems were tasked with assembling information across full text papers to identify salient mechanistic findings (Figure 3A). To be considered a salient finding, a mechanism had to be an experimental result of the study that was mentioned in at least three text passages and/or figure legends in a paper. Mechanistic findings could be simple phosphorylation, binding, and other direct molecular interactions, indirect interactions such as *A increases B phosphorylating C*, or more complex, composite interactions, such as *A when bound to B phosphorylates C*. Changes were made to the index card format to support these changes. A document describing the final index card format can be found in the supplementary material reference in Appendix A.

To measure the ability of the systems to extract salient findings, we needed a set of high-quality, curated findings to compare against. To create this "reference set", three scientists (authors MP, TK and CG) independently curated a set of papers for major mechanistic findings, and then reached consensus on which interactions should be in the reference set. Overall, 56% of the interactions in the reference set were found independently by all three scientists, with the remainder found by two. System output was then compared against the reference set to calculate a recall-like measure called reference set overlap. For reference set overlap, systems were limited to returning a ranked list of up to ten findings per paper.

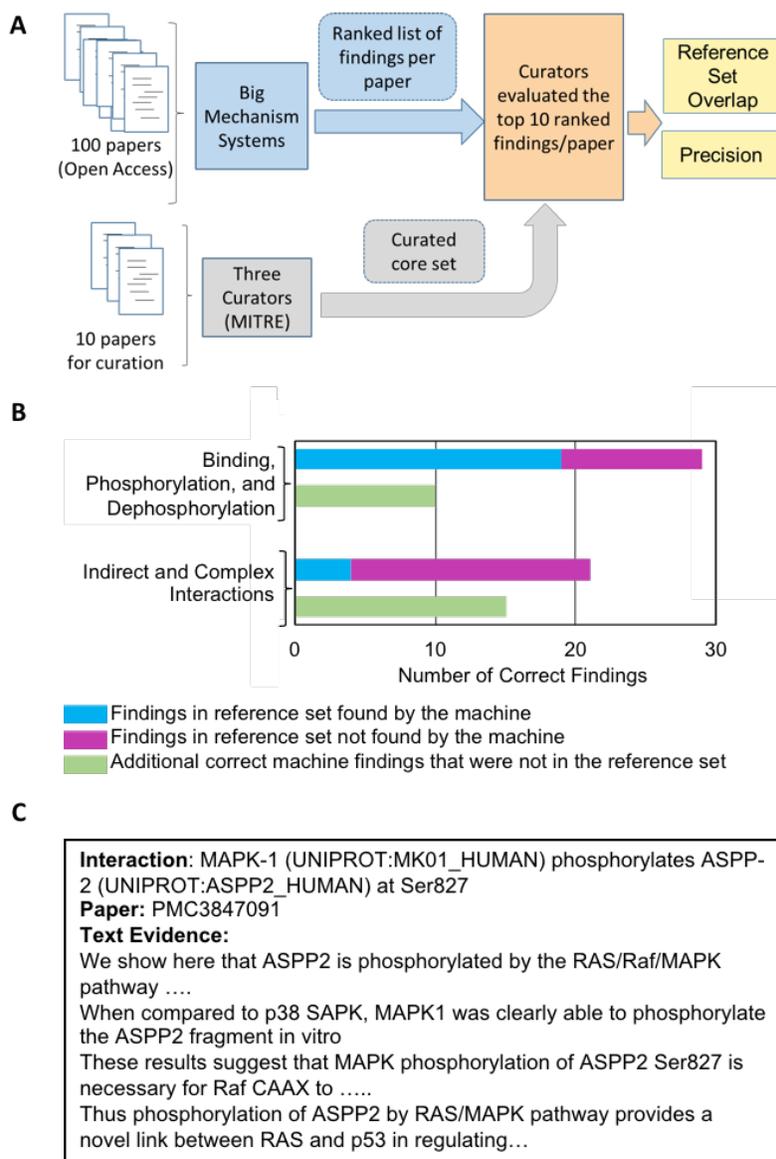

This was done to encourage assembly of similar information (duplicate or near-duplicate interactions got no credit), prioritization of the most important information, and machine assessment of confidence in reading results. We also calculated precision by manual inspection of the evidence for each system's top findings, to give credit for correct extraction of interactions not in the reference set.

Systems successfully identified major findings about phosphorylation and binding. A top performing system identified 66% of the phosphorylation and binding interactions in the reference set with 81% precision (Figure 3B). For phosphorylation and binding interactions found independently by all three curators in the reference set, this system identified 93%. However, retrieval for findings with indirect and complex, composite interactions was low for all systems. In addition, there were errors in assignment of database identifiers to entities; for the system in Figure 3B, among the correct findings, only 81% had both entities correctly grounded to database identifiers. Systems particularly struggled with assigning identifiers for protein families, and 26% identifier failures occurred when there were multiple proteins with the same symbol (Appendix A Table 6). Nevertheless, systems successfully utilized redundancy in the text and assembled information from across a paper (Figure 3C). Overall, the evaluation showed that systems could reliably extract salient findings for two types of molecular interactions that are the backbone of many cancer cells signaling models.

*Figure 3 (A) Phase II evaluation design (B) Results from one submission for reference set overlap (C) An example finding identified by a machine system that utilizes text evidence from across a paper.*

## Phase III: Generating Explanations

For Phase III, the evaluation focus shifted to reasoning. In this evaluation, teams were asked to demonstrate the capabilities of their system to generate *explanations* for biological observations using mechanistic models. The teams were given experimental results to explain and asked to build mechanistic models relevant to particular datasets which included provenance information for each of the interactions in the model (Figure 4A, B). One notable difference between the Phase III evaluations and the previous evaluation phases was that a set of papers was not provided to teams. Instead, teams needed to search the literature or filter large-scale reading results for relevant findings. In additions, teams could also utilize information in curated databases. (Figure 4A, B).

We selected two datasets of changes in protein abundance after drug perturbations to drive the experiments in phase III: one from Korkut et al. [9] (phase IIIa) and one from Fallahi-Sichani at al. [1] (phase IIIb).

Teams were asked to return the following:

- Whether the model could predict the observation, either by simulation or by reasoning over the structure of the model
- For each observation the model explained, the set of interactions involved in the explanation, connecting the perturbation to the measured output (Figure 4C).
- Provenance for all interactions in the models, including text evidence for machine-read interactions.

We then assessed whether each explanation was *plausible* (Figure 4D). To be considered plausible, an explanation had to meet the following criteria:

- The model simulation results or reasoning over the model had to be consistent with the experimental result.
- The explanatory pathway(s) had to fully connect the experimental perturbation to the experimentally measured phosphoprotein.
- Interactions in the explanatory pathway(s) had to be consistent with biological commonsense. For example, if protein A phosphorylates protein B, protein A would need to be a kinase.
- Each interaction in the explanatory pathways must have an information source. If that source was machine reading, then at least one of the text evidence sentences provided needed to support the interaction.
- The explanation had to be consistent with the cell line context. For example, if a gene is knocked out by a mutation in that cell line, it cannot be an active participant in an explanatory pathway.
- The explanation needed to be consistent with explanations for other results, i.e. the same model network needed to be used for all explanations in a cell line.

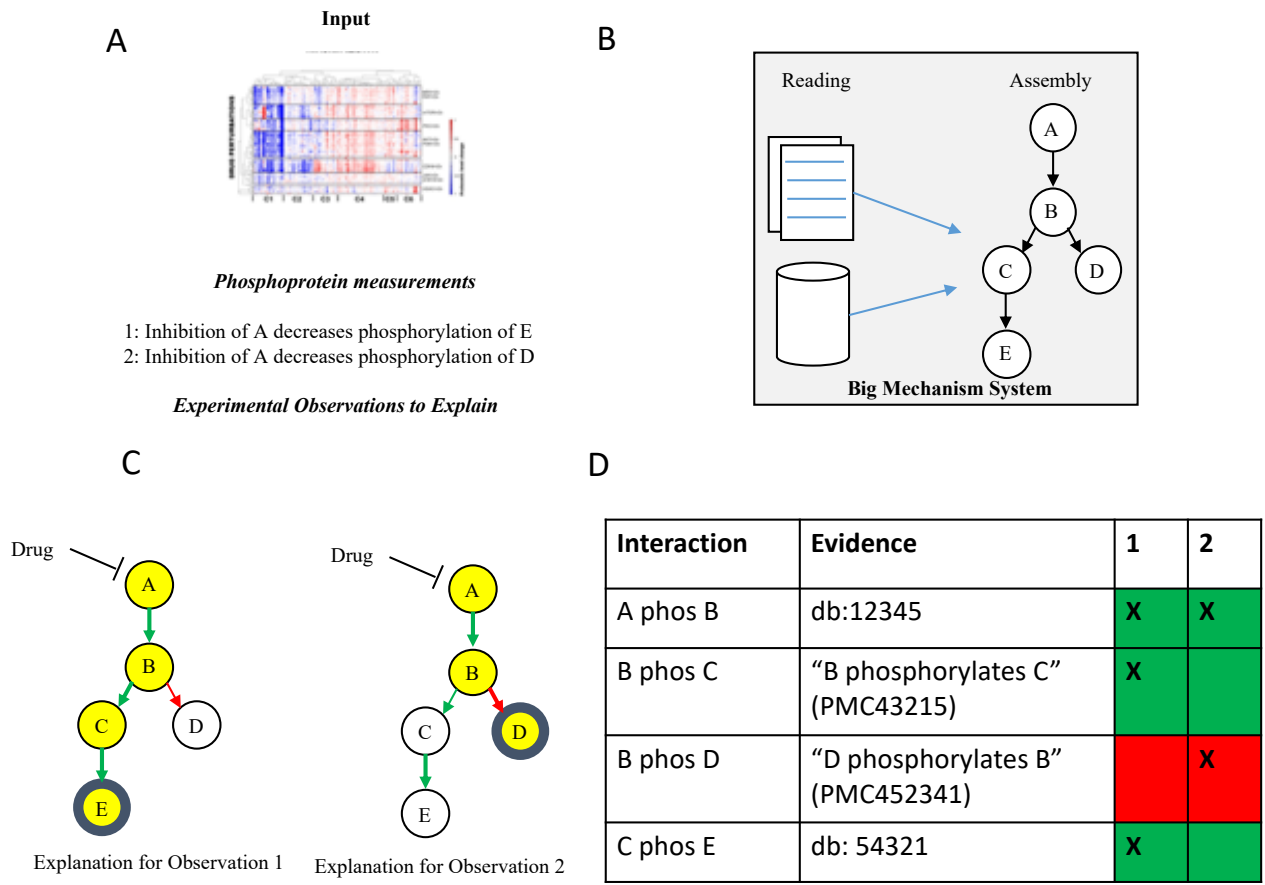

*Figure 4 Phase III Evaluation A) Participants were provided with phosphoprotein data and experimental observations to explain. (B) Big mechanism systems constructed models and parameterized them. (C) Models were reasoned over to generate explanations. Each explanation consisted of a path of interactions from drug target to measured phosphoprotein in an experimental observation. The explanation for observation 1, leading from the inhibition of protein A to the phosphorylation of protein E contains only interactions supported by evidence (green arrows). The explanation for observation two, leading from the inhibition of protein A to the phosphorylation of protein D, contains an unsupported interaction (red arrow). (D) Explanations composed of interactions supported by evidence (e.g. explanation 1) are plausible. Explanations containing interactions not supported by evidence (e.g. explanation 2) are not plausible*

## Phase IIIa

We selected experimental observations for teams to explain from a dataset published by Korkut et al [9]. This paper describes a set of reverse phase protein array (RPPA) experiments that measured the abundance of proteins and phosphoproteins via antibodies (Figure 4A). In this dataset, changes in 143 protein and phosphoprotein abundances were measured in a melanoma cell line both before and after treatment with a set of 89 drugs and drug combinations.  For explanation challenges, we focused on single drug treatments, and identified the largest changes in phosphoprotein abundance in response to drugs. In particular, we selected twenty  the drug-treatment/phosphoprotein abundance pairs where the fold difference between before and after drug treatment exceeded 50% (Appendix A). The evaluation structure for this dataset is described in Figure 4.

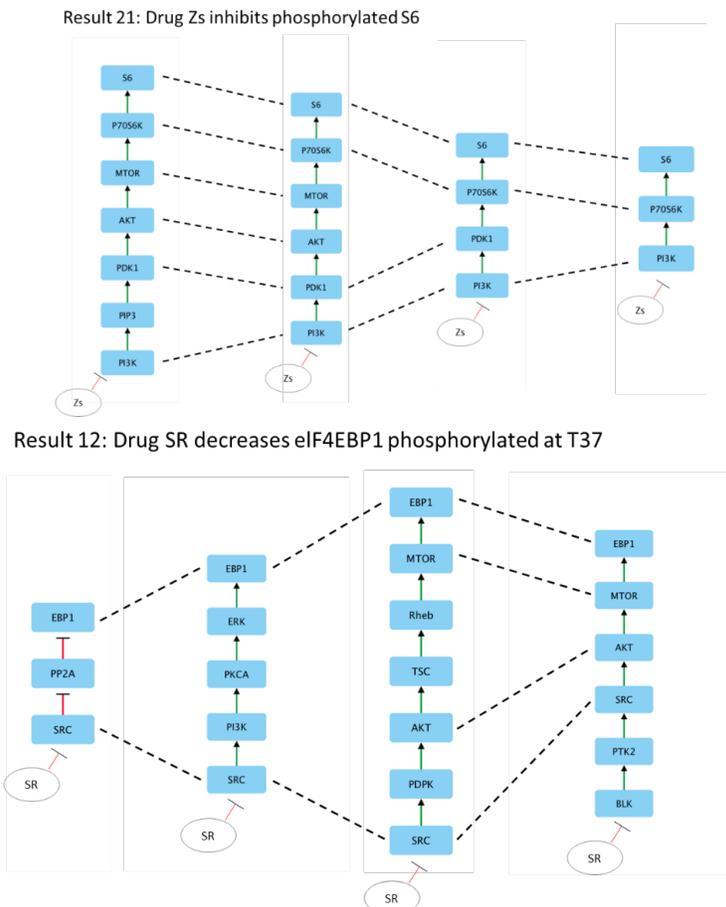

*Figure 5 Explanations submitted for two Korkut experimental results. To facilitate comparisons, explanations are shown in a simplified format. Boxes are labeled with protein name abbreviations. Green arrows indicate increases amount of, activity of, or phosphorylation when phosphorylation activates protein function. Red Ts indicate inhibits, decreases amount of, or dephosphorylates where phosphorylation is known to activate protein function. Each submitted explanation is in a column. Only explanations scored as plausible are shown. Dotted lines mark proteins that are shared among explanations. Explanations for Result 21 were similar to one another, while explanations for Result 12 varied.*

Performer systems collectively provided plausible explanations for all twenty Korkut experimental observations. The top two systems individually generated nineteen and seventeen plausible explanations (Appendix A Figure 8). Machine reading contributed to plausible explanations in both of these systems. In the former, 40% (10/25) of the interactions in explanations were derived from machine reading, and in latter 62% (18/29) of the interactions in explanations were derived from machine reading. For some experimental observations, explanations were consistent among the teams, while for others there were a variety of explanations (Figure 5). Explanations for the top submissions ranged from zero interactions (where the measured phosphoprotein was the drug target) to six interactions long, with median explanation lengths of three and two interactions, respectively.

**Phase IIIb**

After an initial assessment using the Korkut data, we selected a dataset designed to elicit more complex explanations from Big Mechanism systems. The dataset from Fallahi-Sichani et al. [1] describes the response of phosphoproteins to drug perturbations across a wide range of cell lines, drug concentrations, and across a longer time course with earlier time points than those in the Korkut dataset. Six of the cell lines in this study were associated with both mutation data and baseline gene expression data.

Explanations for findings in the Fallahi-Sichani data set had the potential to be much more complex than those from the Korkut data set because we asked teams to explain observations that involved changes between cell lines with different mutation profiles. As an example, two cell lines with different mutation profiles may respond differently to a drug. An explanation for the difference in the response of a particular protein to a drug may involve comparing the differential function of a particular pathway between those two cell lines (Figure 4). For the purposes of the evaluation on the Fallahi-Sichani dataset, we generated a set of five findings that captured interesting differences in behavior in response to drug perturbations across the cell lines studied. Two these (Finding #1 and Finding #5) are shown below, while the full set of findings is described in Appendix A.

- We observe a dose-dependent decrease of p-S6(S235/236) across five cell lines and drugs. Feel free to use any or all of the time points in your explanation.
- We observe a dose-dependent increase in total c-Jun in the cell line RVH421 but a dose-dependent decrease in total c-Jun in the cell line C32.

We asked teams to attempt to explain the first of five findings in addition to at least one of the other four findings. Each team returned a set of explanations for each observation that they attempted and a comparison of the results of the model compared to the results presented in the Fallahi-Sichani paper. Collectively, the teams generated explanations for all five findings.

For Finding #1, collectively teams proposed two possible mechanisms for the change in phosphorylation – one going through P90 S6 Kinase, and the other going through P70 S6 kinase. Some groups identified only one of these two pathways, while others identified both, and justified the choice of one based on expression data or other information about the cell line. This is an example of Big Mechanism systems identifying competing hypotheses that could be tested in a laboratory.

The explanations teams submitted for Findings #2 through #5 varied more substantially than those for the Korkut dataset, as expected, given that the findings involved multiple cells lines. For example, for Finding #5, one group highlighted potential bistability caused by the network structure as a potential cause of the difference in c-Jun expression, while another used expression data to explain the observation by highlighting differential expression of two upstream signaling pathways that result in c-Jun expression, one of which is not affected by the treatments.

## Summary of Performer Approaches

In addition to determining whether Big Mechanism systems could accomplish these tasks, an important part of the evaluation was to learn what methods worked for doing so. To gather this information, we asked teams to send us information about their methods.

Teams took a variety of approaches in applying machine reading and selecting interactions for their models. Some groups used a single machine reading technology, while others used multiple reading systems and combined the output. Some groups used queries to identify relevant papers to which they then applied machine reading, while others read the entire PubMed Central Open Access subset. This resulted in varying numbers of interactions extracted, ranging from thousands to millions of machine-read interactions per submission. To select interactions for their models, all teams winnowed down the machine-read interactions, typically reducing the set of interactions by three orders of magnitude in selecting interactions for their models. The resulting models included 10s to 1000s of interactions. A variety of approaches were taken to filter and assemble interactions, including combining similar interactions, selecting interactions connected to entities in the experimental data, filtering interactions based on the biological properties, requiring a certain amount of text evidence, and using human review to remove incorrectly read interactions.

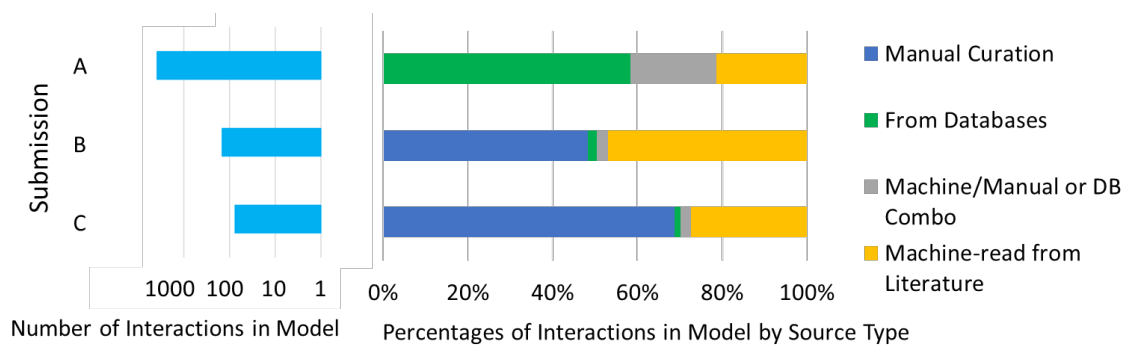

*Figure 6 Percentages of interactions from machine reading, curated databases, de novo manual curation, and combinations of these in models built for explaining Fallahi results. Only systems that provided a plausible explanation for Fallahi result 1 using a de novo*

Nearly all teams chose to combine machine reading with information manually curated for this task and/or information from public, curated databases. Two teams started with a manually curated model and extended it with machine reading, as, for example, in, [10]. Three other teams combined machine reading with manual curation and/or public database information to build models *de novo* – see, for example, [11]. In submissions for the Phase IIIb task for these three teams, 21% to 47% of the interactions in their models came exclusively from machine reading (Figure 6). This shows that human curation plays an important role in Big Mechanism models at this time.

The performers in the Big Mechanism program applied different modeling and analysis methodologies to this problem. Models were represented by Boolean networks, Petri nets[12], as well as rule-based representations[13]. Some teams parameterized and simulated a model, while others used static analysis methods to generate explanatory paths in the model[12], [14]. Model building approaches also differed, including building a model *de novo* by combining interactions from structured databases and reading results[12], [14], as well as iteratively building on existing curated models[10].

In addition, all teams incorporated human expertise into their workflows, but did so in varying ways. For example, some teams augmented an expert-curated base model with reading results, while others combined human-curated databases with reading results in automated assembly pipelines. Humans also played a role in the filtering steps taken by teams. Some teams manually filtered results produced by the system, while others used expert insight to build biological "common sense" filters (e.g., phosphorylation reactions can only be catalyzed by kinases). These role of humans in the Phase III experiments is summarized in Table 1.

*Table 1 The role of humans in the Korkut and Fallahi experiments*

|  | Submission | | | | | |
|---|---|---|---|---|---|---|
|  | 1 | 2 | 3 | 4 | 5 | 6 |
| Human curation of interactions for the base model (i.e. includes interactions that were not suggested by machine reading)? | Yes | Yes | No | Yes | No | No |
| Used information from human-curated databases? | Yes | Yes | No | No | Yes | Yes |
| Human input to bio commonsense into automated filters? | No | Yes | No | No | Yes | Yes |
| Human scoring of machine-read interactions to get error rates used in filters? | No | No | No | No | No | Yes |
| Human filtering of machine-read interactions? | No | Yes | Yes | Yes | No | No |
| In-depth human review and corroboration of machine read interactions for selection for model? | No | Yes | No | No | No | No |
| Human selection of modeling methods for specific tasks? | No | No | No | No | No | Yes |
| Human parameterization of models? | No | No | No | Yes | No | No |
| Human identification of explanation pathways? | No | Yes | Yes | Yes | No | No |

## Performer Feedback on the Evaluations

Because the evaluation structure for the program had a unique structure relative to biological text mining and systems biology evaluations, the evaluation team collected feedback at the end of the program from performers through interviews. Interview questions focused on the following areas:

- **How did the evaluation affect research?** What did you do differently due to the evaluation? What avenues were shut off? What did you learn?

- **What are your thoughts on the evaluation methods?** Was the iterative approach helpful in accomplishing goals? Was the pace of iteration appropriate? What was the impact of evaluations on program progress?

- **Did your system make contributions to scientific research?** Did you contribute to biological knowledgebases, generate testable hypotheses about cancer signaling, or help better understand biological pathways?

## Evaluation's Effect on Research

One of the major themes of the interview responses was that the evaluations revealed the state of the technologies being developed and identified gaps. The Phase I evaluation identified a need for the assembly of information within and across papers, and the Phase II evaluation highlighted grounding as a bottleneck for integrating reading into systems. Because systems were not required to be fully-automated, performers were able to identify parts of the system that could be automated over time. Teams reported that they pursued a number of research avenues that were direct results of the evaluation, including the creation of automated assembly frameworks, the creation of "biological common sense" systems to work with "noisy" automated reading results, and the creation of technologies for automated explanation generation. At the same time, performers noted that initially proposed research avenues were not pursued, as they were not relevant to the evaluation structure. One example brought up by multiple teams was reading for experimental approach and observations. Teams also noted that while the technology built could be the underpinnings of a system for use by biologists, the systems developed did not focus on the potential end users.

## Performer Opinions of Evaluation

The teams provided a great deal of candid feedback on the evaluation structure and approach. Overall, the iterative and collaborative approach was well received. Performers appreciated the dry runs to resolve technical issues, and also the calls for their input during the evaluation design process. Performers noted that the common tasks helped drive progress, but that the datasets weren't ideal for all of the teams' modeling approaches, which limited their utility. Similarly, the common data formats were seen as helpful to drive collaboration, but some teams noted that this steered the reading efforts and limited the types of information that were being extracted. Finally, some teams would have preferred to see modeling and explanation evaluated earlier in the program.

The area where performers saw the largest contribution due to evaluation activities was in the acceleration of qualitative and semi-quantitative model building and analysis. However, as mentioned above, some aspects of the evaluation limited the ability to answer scientific questions – the Phase IIIa dataset was limiting due to the number of timepoints captured, and the Phase IIIb dataset was limiting due to the number of phosphoproteins measured.

# Lessons Learned about Evaluation of Complex Systems

## Evaluation in the Absence of a Gold Standard

While we initially set out to develop and use gold standards for at least part of the evaluations, we ultimately took the approach of evaluating based on evidence, expert review, and plausibility. One of the original goals of the Phase I evaluation was to use expert human curation to generate a gold standard set. However, it became immediately clear that due to the complexity of the task, it would require extensive training of curators and significant time on their part to produce a dataset of sufficient quality. Instead, we moved to an evidence-based evaluation, where teams were asked to "show their work" to enable semi-automated evaluation by domain experts.

In Phase II, to encourage systems to move towards extracting the main novel mechanistic findings from each paper, we designed a new evaluation based on a small "reference set" of interactions carefully curated to represent major mechanistic findings of each paper. This allowed us to calculate basic measures such as reference set overlap (a proxy for recall) and precision for the systems.

Phase III presented an opportunity for automated comparison between modeling output and experimental evidence in a manner similar to community challenges such as the DREAM challenges [6]. However, concordance between model predictions and experimental results did not specifically assess performance in the area that the Big Mechanism program was focusing on – namely the explanatory power of mechanistic models. This led to the development of an evaluation which focused on explanation, rather than prediction, based on system output that consisted of sets of explanatory interactions backed by provenance and evidence information.

## Incremental and Iterative Approach and Evaluation as a Partnership

The evaluation phases were incremental, enabling components and capabilities to be added as they came on line. Within each evaluation phase, there were multiple dry runs before the final evaluation. This allowed performers to work through technical difficulties and iteratively improve their systems, and enabled evaluators to refine the submission format and improve their methods for evaluating system output. In addition, evaluation tasks were common across teams, even though the teams had disparate goals and capabilities. This enhanced collaboration among performer teams and the evaluators. Collaboration was further enhanced by the evaluations' use of common representations with a minimum information format, which helped drive data sharing and interoperability. Finally, each evaluation phase required that performers show their work by presenting evidence and/or explanations for their system outcomes. While this was necessary for evaluators to assess results, especially in the absence of a gold standard, this information was also critical for building trust in a system for potential end users.

## Common Representations and "Minimum Information" Standards for Data Sharing

The Big Mechanism program supported the development of multiple modeling approaches, each with their own information needs and representations. Some approaches required detailed information about interactions and the molecules involved, while others only required basic information about the interaction. In the Big Mechanism program, the partnerships and collaborations across performers have evolved with the program; as new technologies were developed, they were integrated into different end-to-end systems. To facilitate this, we studied the various representations in use by the teams and developed a "neutral" JSON format with a set of minimum information requirements that could be easily extended by reading and assembly systems. This was designed to maximize the amount of information sharing and technology exchange. A detailed description and examples of the Index Card format can be found in the Git repository linked to in Appendix A.

## Discussion

The evaluations demonstrated that several Big Mechanism goals were achieved: systems were built to read at scale (Figure 2, Figure 6). Semi-automated Big Mechanism systems were able to assemble mechanistic fragments from literature into mechanistic models (Figure 3). Furthermore, semi-automated Big Mechanism systems were able to generate plausible explanations for experimental results (Figure 5). Finally, components of Big Mechanism systems were able to compensate for machine reading errors; even though machine reading did not achieve the accuracy of human curation, machine reading nevertheless contributed to the building of models and plausible explanations. The ability to use imperfect machine reading for generating plausible explanations was achieved through three general means: incorporating human experts in the loop within Big Mechanism systems, exploiting expert curated resources as additional knowledge sources, and making use of redundancy.

From the beginning, the Big Mechanism program did not mandate a fully-automated approach, but instead encouraged performers to explore how humans could be integral, productive components of Big Mechanism systems. Teams responded by incorporating human experts into their pipeline in various ways. These included manually curating base models that were extended by machine reading, providing feedback to automated components, and manually filtering and selecting interactions and explanations. Another role where humans were critical was in framing and devising ways to tackle diverse problems, particularly given the diversity of problems in the Phase IIIb task. In addition, while some teams achieved automated generation of explanations, human experts were needed to understand and interpret them in a larger context.

Human expert knowledge was also a critical feature in the Big Mechanism automated components. Domain expert knowledge was hardcoded in machine reading rules, in rules using biological commonsense for automated filtering, and in rules for finding and assessing explanations. Incorporating human domain expertise was particularly important in improving grounding to database identifiers. When evaluations showed that machines made errors in grounding complexes and protein families, one team manually developed a resource of standard identifiers for protein families and complexes, and used this to improve accuracy [15]. Expert human curation was also a critical contributor to models and explanations. For models that explained the majority of experimental results, greater than half the interactions were derived from human curation, either from *de novo* curation or from expert curated publicly available databases. This may reflect current but surmountable limitations in machine reading, and may also be a consequence that only a fraction of papers are open access and available to machine reading.

Automated Big Mechanism system components also compensated for machine reading errors by leveraging redundancy. The Phase I evaluation showed that machine reading generated a great deal of redundant information; the Phase II evaluation was structured to challenge systems to automatedly assemble redundant information within papers. Teams took this further, utilizing redundancy across different levels, including across papers, across sources (e.g. papers and curated databases), and across different levels of generality, such as interactions that involve a specific protein versus multiple members of a protein family. Teams used redundancy as a way to ascertain how likely an automatically extracted interaction was to be accurate, assigning increased confidence to interactions with multiple supporting pieces of evidence.

The evaluations and feedback suggest a few directions for future work on Big Mechanism systems. First, scientists need not just assertions about mechanisms, but also want information about experimental evidence. While some performers were initially interested in reading for experimental evidence, this was not an explicit element of the evaluations. However, given the importance of this information, it would be useful to do further research on the extraction and assembly of experimental evidence, and how to provide it in a format that is readily usable by scientists. Second, hypotheses and explanations generated by Big Mechanism systems should be tested in the laboratory to evaluate the potential contribution of Big Mechanism systems to science. Finally, future work will test whether Big Mechanism systems can be useful for other biological problems and even for other non-biological domains.

## Acknowledgements

This work was produced for the U. S. Government under Basic Contract No. W56KGU-18-D-0004-0001 and is subject to the Rights in Technical Data Noncommercial Items clause at DFARS 252.227-7013 (FEB 2012).

We gratefully acknowledge the critical contributions of Dr. Paul Cohen, the original DARPA Program Manager and originator of the Big Mechanism Program. He not only had the vision for the program but was instrumental in shaping the framework for the evaluation. We also acknowledge the valuable guidance provided by Mr. Stephen Jameson in taking the program to completion and building on its lessons learned. In addition, we acknowledge Mr. William Bartko, who provided invaluable program support and feedback and last but not least, to the Big Mechanism performers, who have made amazing progress towards creating Big Mechanism systems.

# References/Bibliography

# Appendix A

## Phase I Evaluation

### Index card details for the Phase I evaluation

**Index Card Metadata**
- PMC ID of Paper
- Timestamps
- Source (entity generating the Index Card)
- Type of Source (Human/Machine/Human-Machine)

**Relationship to Model:**
- **Type:** Indicate one of these four types: Extension; Specification; Corroboration; Conflicting
- **Model Element:** For Specification, Corroboration, or Conflicting, indicate what is referenced in the model, using its identifier; for Extensions, there will be no reaction in the model by definition.
- **Model Participants:** Entities in the extracted information that are also in the model. In JSON format, this information is a subfield of Participant.

**Extracted Information:**

- **Participant A**: A protein, gene or chemical entity, or a complex of entities. If it is a left to right interaction, this is the left/subject entity. This includes the following subfields:
    - Grounded Entity Id (e.g. UniProt id);
    - Entity Type: Protein, Chemical, or Gene;
    - Features (as applicable): Specify any protein modifications, isoform, or mutant form;
    - Whether entity is in the model (values are Yes/No)

- **Participant B**: A protein, gene or chemical entity, or a complex of entities. If it is a left to right interaction, this is the right/object entity. This has the same subfields as Participant A.

- **Interaction Type**: Indicate one of the following, and complete any relevant corresponding subfields:
    - 1) Binds
        - Subfield: Binding site
    - 2) Modifies or Inhibits Modification
        - Subfield: Modification (Type and Position)
    - 3) Translocates

- Subfields: From and To, using the Gene Ontology subcellular localization as potential value
    - 4) Increases or decreases
    - 5) Increases or decreases activity

- **Negative Information** Indicate whether the interaction occurs, or whether this is "negative" information, and the evidence indicates this interaction does not occur; values are Yes/No.

**Evidence:**
One or more text passages, tables, or figures from the paper supporting the extracted information. The evidence must be based on experimental results from this study, and can be a sentence or short passage from Results, a figure legend, a table (provide the table number and name), or an assertion based on this paper's results from any section of the paper. Any single, contiguous piece of evidence should not be more than a few sentences long. More than one piece of evidence can be submitted to support an extracted piece of information. However, only one piece of evidence is required for each interaction, as long as it contains sufficient evidence to support all the annotated information in the index card.

## Methods for scoring index cards in the Phase I evaluation

We manually evaluated index cards to assess whether they were "largely correct" or incorrect. Scored cards were assessed as largely correct when they met the following criteria:

- The interaction on the card was consistent with the text evidence.

- Both participants were correct, as well as the interaction type, given the text evidence. Putting a blank for one participant was allowed as largely correct.

- The negative information indicator, TRUE/FALSE, was consistent with the text evidence.

- Correctness of entity grounding was not a factor in calculating precision. Having a correct entity text string was sufficient.

- Inclusion and correctness of phosphorylation sites was also not a factor in calculating precision.

In addition, some index cards were skipped and not scored as either largely correct or as incorrect. For the precision calculation, these skipped cards were not counted in the denominator. Index cards were skipped when the following occurred:
- The text evidence was about background or methods rather than experimental results of the paper.
- The interaction on a card was a repeat of an interaction already scored on another card submitted by the same team for the same paper.
- The interaction was of the form: <blank> increases (or decreases) amount of a protein. These were skipped because they contain no useful information.

Index cards that were not skipped and failed one or more criteria for being largely correct were scored as incorrect.

## Index cards reviewed in the Phase I evaluation

We reviewed all the cards from 8 to 13 papers per team submission. The number of papers varied among submissions because the number of index cards generated varied greatly among teams, and for submissions with fewer cards, we needed to score cards from enough papers to get reasonable precision numbers. Consequently, the precision numbers are based on a different composition of papers for each team. The papers we scored cards from for each team submission are:

*Table 2 Scored Cards by Team*

| PubMed Central ID | Submission 1 | Submission 2 | Submission 3 | Submission 4 |
|---|---|---|---|---|
| PMC2212462 | scored | scored | scored | scored |
| PMC1240052 | scored | scored | scored | scored |
| PMC1392235 | scored | scored | scored | scored |
| PMC2156209 | scored | scored | scored | scored |
| PMC2171478 | scored | scored | scored | scored |
| PMC2172453 | scored | scored | scored | scored |
| PMC2193139 | scored | scored | scored | scored |
| PMC3284553 | scored | scored | scored | scored |
| PMC2172734 | scored |  | scored | scored |
| PMC4122675 | scored |  | scored | scored |
| PMC2442201 |  |  | scored |  |
| PMC2585478 |  |  | scored |  |
| PMC3102680 |  |  | scored |  |

## Metrics calculations for Phase I

Precision was calculated as:

$$\frac{\text{\# largely correct cards}}{\text{\# largely correct} + \text{\#incorrect cards}}$$

To calculate an estimate of the number of correct cards generated per day, we first estimated the fraction of the submitted cards that were largely correct, based on the cards we scored:

$$\text{Estimated \# largely correct} = \frac{\text{\# largely correct cards}}{\text{\# largely correct} + \text{\#incorrect} + \text{\#skipped cards}}$$

To estimate the number per day, we divided the estimated number of largely correct cards by the number of days given for machine-only and for human + machine conditions. For machine-only, this was seven days. For human + machine, this was three days.

## Results on precision and correct cards per day for Phase I

The table below shows the numbers corresponding to results in D:

*Table 3 Results on precision and correct cards per day for Phase I*

| Submitter-Condition | Precision | Largely Correct Cards per Day (estimated) |
|---|---|---|
| Submission 1-Machine | 0.59 | 62 |
| Submission 1 Human+Machine | 0.96 | 8 |
| Submission 2-Machine | 0.23 | 342 |
| Submission 2 Human+Machine | 0.67 | 25 |
| Submission 3-Machine | 0.63 | 110 |
| Submission 3 Human+Machine | 0.95 | 24 |
| Submission 4-Machine | 0.49 | 695 |
| Submission 4 Human+Machine | 0.58 | 215 |

# Phase II Evaluation

## Index card details for the Phase II evaluation

## Materials and methods for the Phase II evaluation

We selected papers about molecular interactions related to the Ras-signaling pathway from the PubMed Central open access set. The PubMed Central ids for the 10 papers in the test set are provided in Table 4. In addition, the teams were provided with a small model, which is available at https://gitlab.mitre.org/Big-Mechanism/tech-report.

*Table 4 Phase II PubMed Central IDs*

| PubMed Central ID |
|---|
| PMC1234335 |
| PMC3178447 |
| PMC3690480 |
| PMC4345513 |
| PMC534114 |
| PMC2841635 |
| PMC3595493 |
| PMC4052680 |
| PMC4329006 |
| PMC4729484 |

For assessing precision, we manually scored interactions in Phase II as was done in Phase I with one exception. Unlike in Phase I, in Phase II cards that were not about results and were incorrect were scored as incorrect. As in Phase I, in Phase II, cards that were not about results and were largely correct were skipped.

To create a reference set of interactions, three curators with expertise in biology (authors MP, CG, and TK) independently read and identified mechanistic interactions from these papers using the following criteria:

The interaction was one of:

- 1) a direct mechanistic interaction of the following types: phosphorylation, binding, dephosphorylation, translocation, increases/decreases amount of;
- 2) a complex interaction, involving two or more of the above interactions, such *as A when bound to B phosphorylates C*, or
- 3) a complex indirect interaction where an entity increases an embedded interaction, such as *A increases B phosphorylating C*.

The mechanistic interaction had to be an experimentally supported finding of the paper, and not background information.

The interaction had to be mentioned in at least three text passages and/or figure legends in a paper.

After working independently, the three curators met to agree on a final list for the reference set, based on accuracy and the criteria above. Only interactions that were found independently by at least two curators were included in the final reference set. When complex interactions were included in the reference set, the embedded component interactions were also included as a separate interactions in the reference set, in addition to the full complex interaction. The reference set interactions for the ten test set papers and the dry runs are available at https://gitlab.mitre.org/Big-Mechanism/tech-report/tree/master/data/phase_2/reference_set.

To score reference set overlap, we evaluated whether there was a matching interaction for each interaction in the reference set in the top ten ranked interactions submitted by each team. To be a match, the submitted interaction had to include both participants in the reference set interaction and the same interaction type; a generic participant was only permitted if the reference set interaction had a generic participant. Interactions that were similar to the meaning of a reference set interaction, but captured in a somewhat different way that was also accurate, were accepted as matches. In order to be a match, an interaction had to also be scored as largely correct under the precision scoring criteria (e.g., the text evidence had to support the interaction; grounding errors were permitted.)

To make scoring more efficient, we wrote a program to automatically assess and flag matches. These were then reviewed manually. Interactions that were not flagged were also were also manually reviewed to identify matches to additional reference set interactions, particularly for interactions that were similar in meaning to a reference set interaction but were inexact matches.

### Phase II Evaluation Results

Table 5 below shows the Phase II results for reference set overlap and precision for the test set of ten papers.

*Table 5 Reference set overlap and precision for four reading and assembly system submissions[1].*

| | Submission | | | |
|---|---|---|---|---|
| **Statistic** | **1[1]** | **2** | **3** | **4** |
| Number of matches to phosphorylates, dephosphorylates, and binds interactions in the reference set (Reference set contained 29 such interactions) | 19 | 22 | 16 | 18 |
|    Percent Reference Set Overlap | 66 | 76 | 55 | 62 |
| | | | | |
| Number of matches to indirect and complex interactions in the reference set (Reference set contained 21 such interactions) | 4 | 0 | 0 | 0 |
|    Percent Reference Set Overlap | 19 | 0 | 0 | 0 |
| | | | | |
| Precision (all interaction types) | 0.74 | 0.63 | 0.46 | 0.45 |

[1]Submission 1 corresponds to the results shown in Figure 3B.

Table 6 below shows the types of errors observed in "grounding:" associating names and symbols of biological entities with unique identifiers from reference databases such as UniProt for proteins.

*Table 6 Errors Observed in Grounding to Database Identifiers in Phase II*

| Entity Text | Number of Submissions with this Error | Error Description | Error Type |
|---|---|---|---|
| pTpYERK | 1 | not grounded | symbol for modified form of a protein |
| ERK | 4 | 3 submitters grounded to protein EPHB2; 1 submitter did not ground | protein family |
| RAS | 1 | not grounded | protein family |
| SAPK | 1 | grounded to MAPK9; full text on this is "p38 SAPK" ; there are 4 MAPK proteins that are p38 SAPKs. MAPK9 is not one of them. | protein family/not capturing full protein name/synonym issue |
| histone | 1 | not grounded | complex of proteins (not part of model) |
| Extracellular Signal-Regulated Kinase | 1 | grounded to protein EPHB2 | protein family |

| PKA | 2 | in one case grounded to a protein in frog; another case, not grounded. PKA refers to a family of proteins | protein family; also wrong species |
| --- | --- | --- | --- |
| PKD1 | 1 | grounded to PKD1, another human protein with same synonym | synonym issue |
| ISO | 1 | grounded to a protein, instead of chemical | chemical vs protein identification; synonym issues |
| MEK | 1 | not grounded | protein family |
| SOS | 1 | not grounded- member of complex not grounded | protein family |
| p85 | 2 | grounded to ARHG7, another human protein with same synonym; grounded to PPP1R13B, another human protein with the same synonym; missing identifier | synonym issue |
| 14-3-3zeta | 1 | grounded to wrong species (mosquito) | wrong species |
| RTK | 1 | Not grounded. this refers to a family of proteins. The paper uses it sometimes as a stand-in for FGFR2, an RTK that they do experiments in | protein family/synonym issue |
| HSP20 | 1 | grounded to wrong heat shock protein (HSP22) | incorrect protein-reasons unclear |
| PI3K | 1 | grounded to a protein named PI3 instead of with proteins from the PI3K complex | protein complex/ synonym |
| MEK | 1 | not grounded | protein family |
| AKT | 1 | not grounded | protein family |
| Pi3K p110 | 1 | grounded to wrong subunit | incorrect protein- involves protein family issue |

# Phase III Evaluation

## Phase IIIa

In order to assess the ability of the DARPA Big Mechanism performer systems to make accurate predictions and generate explanations for observed experimental results, we assembled a list of highly-changed phosphoprotein abundances from a dataset published by Korkut et al. [9]. The data were collected from the melanoma cell line SkMel-133 and consisted of 89 single drug or drug combination treatments. Measurements of protein/phosphoprotein abundance were made for 138 entities using Reverse Phase Protein Array (RPPA). In addition, the data set includes cell viability measurements. We selected a subset of these observations as target for explanation based on the following criteria:

1) The experiment was a single drug perturbation
2) The fold change in protein abundance was >1.5x or <0.5x

*Table 7 Phase III Evaluation: Korkut results to explain*

| Experimental Observation Number[1] | Treatment Description (drug abbreviation \| dose or time-point) | Drug Target | Antibody ID | Fold Change[2] |
|---|---|---|---|---|
| 1 | 901\|3 | MEK1/2 | MAPK_pT202 | 0.468 |
| 2 | AK\|10 | AKT | AKT_pS473_V | 0.140 |
| 3 | AK\|10 | AKT | AKT_pT308_V | 0.241 |
| 4 | AK\|5 | AKT | GSK3a_b_pS21 | 0.440 |
| 5 | AK\|10 | AKT | S6_pS235_V | 0.274 |
| 6 | AK\|10 | AKT | S6_pS240_V | 0.418 |
| 7 | RO\|7 | PKC | GSK3a_b_pS21 | 1.586 |
| 8 | RO\|7 | PKC | S6_pS235_V | 0.301 |
| 9 | RO\|7 | PKC | S6_pS240_V | 0.468 |
| 12 | SR\|2.4 | SRC | 4EBP1_pT37_V | 0.442 |
| 13 | SR\|4.8 | SRC | CHK2_pT68 | 1.749 |
| 14 | Tm\|0.6 | mTOR | AKT_pS473_V | 3.187 |
| 15 | Tm\|0.6 | mTOR | AKT_pT308_V | 2.187 |
| 16 | Tm\|0.6 | mTOR | p70S6K_pT389_V | 0.331 |
| 17 | Tm\|0.6 | mTOR | S6_pS235_V | 0.058 |
| 18 | Tm\|0.6 | mTOR | S6_pS240_V | 0.050 |
| 19 | ZS\|1.2 | PI3K | AKT_pS473_V | 0.204 |
| 20 | ZS\|1.2 | PI3K | p70S6K_pT389_V | 0.495 |
| 21 | ZS\|1.2 | PI3K | S6_pS235_V | 0.273 |
| 22 | ZS\|1.2 | PI3K | S6_pS240_V | 0.442 |

[1] Originally there were 22 results to explain; two were removed because the drugs have multiple off-target effects.

[2] Phosphoprotein abundances that increased are marked in pink and phosphoprotein abundances that decreased are marked in green.

Additional information about the experiments and data are available in Korkut et al. [9], and at the supplementary website for that paper (http://www.sanderlab.org/pertbio/).

| Submission | Observation 1 | Observation 2 | Observation 3 | Observation 4 | Observation 5 | Observation 6 | Observation 7 | Observation 8 | Observation 9 | Observation 12 | Observation 13 | Observation 14 | Observation 15 | Observation 16 | Observation 17 | Observation 18 | Observation 19 | Observation 20 | Observation 21 | Observation 22 | Number Supported |
|---|---|---|---|---|---|---|---|---|---|---|---|---|---|---|---|---|---|---|---|---|---|
| 2 | S | S | S | S | S | S | S | S | S | N | S | S | S | S | S | S | S | S | S | S | 19 |
| 6 | S | S | S | S | S | S | N | S | S | S | S | N | N | S | S | S | S | S | S | S | 17 |
| 4 | S | S |   | S | S |   | S | S |   | S | I | S |   | S | S |   | S | S | S |   | 13 |
| 1 | S | S | S | N | S | S | I | N | N | N | N |   | S | S | S | S | S | S | N | N | 11 |
| 5 | S | S | S | S | S | S |   |   | S |   | N | N | N | N | S | S | S | S |   |   | 11 |
| 3 | S | S |   | N | N |   | S | N |   | N |   | N | S |   | N | N | N |   |   |   | 4 |

*Figure 7 Explanation results for teams for 20 Korkut experimental observations. Green indicates explanations scored as plausible. Collectively, teams found plausible explanations for all 20 observations. Green squares indicate an explanation supported by evidence, purple squares indicate an explanation not supported by evidence, brown squares indicate an incorrect prediction by the model, and white squares indicate an explanation that was not attempted or a prediction that was out of scope of the model.*

## Phase IIIb

We provided five observations from a dataset from Fallahi-Sichani et al. [1] as targets for explanation by Big Mechanism performer teams. We required that teams explain the first of these observations and at least one of the four remaining observations. These observations are summarized in Table 8 below).

Table 8. In addition to descriptions of each finding, we also provided figures generated from the Fallahi-Sichani et al. dataset. The figure that we provided for Finding 1 is replicated here (Figure 8 below).

*Table 8 Fallahi Findings to Explain*

| Finding ID | Cell Lines | Drugs | Antibody Target | Finding |
|---|---|---|---|---|
| 1 | C32 LOXIMVI MMACSF MZ7MEL RVH421 | AZ628 PLX4720 SB590885 vemurafenib Selumetinib | p-S6(S235/S236) | We observe a dose-dependent decrease of p-S (S235/236) across five cell lines and drugs. Feel free to use any or all of the time points in your explanation. |
| 2 | LOXIMVI RVH421 | AZ628 PLX4720 SB590885 vemurafenib Selumetinib | perk(T202/Y204) | We observe an increase in the abundance of pERK(T202/Y204) at low dosages across all drugs in the LOXIMVI cell line at 1 hour after treatment. In contrast, the cell line RVH421 shows a dose-dependent decrease in the abundance of pERK(T202/Y204) across the time course. |
| 3 | LOXIMVI | AZ628 | p-Histone H3(S10) | We observe a dose-dependent decrease in the abundance of p-Histone H3(S10) in the cell line LOXIMVI in response to AZ628 but not in response to the other drugs. |
| 4 | WM115 C32 | AZ628 PLX4720 SB590885 vemurafenib | p-AKT(S473) | During intermediate time points, we observe a decrease in the abundance of p-AKT(S473) in the cell line WM115 but not in the cell line C32. |

| 5 | RVH421 C32 | AZ628 PLX4720 SB590885 vemurafenib Selumetinib | Total c-Jun | We observe a dose-dependent increase in total c-Jun in the cell line RVH421 but a dose-dependent decrease in total c-Jun in the cell line C32. |

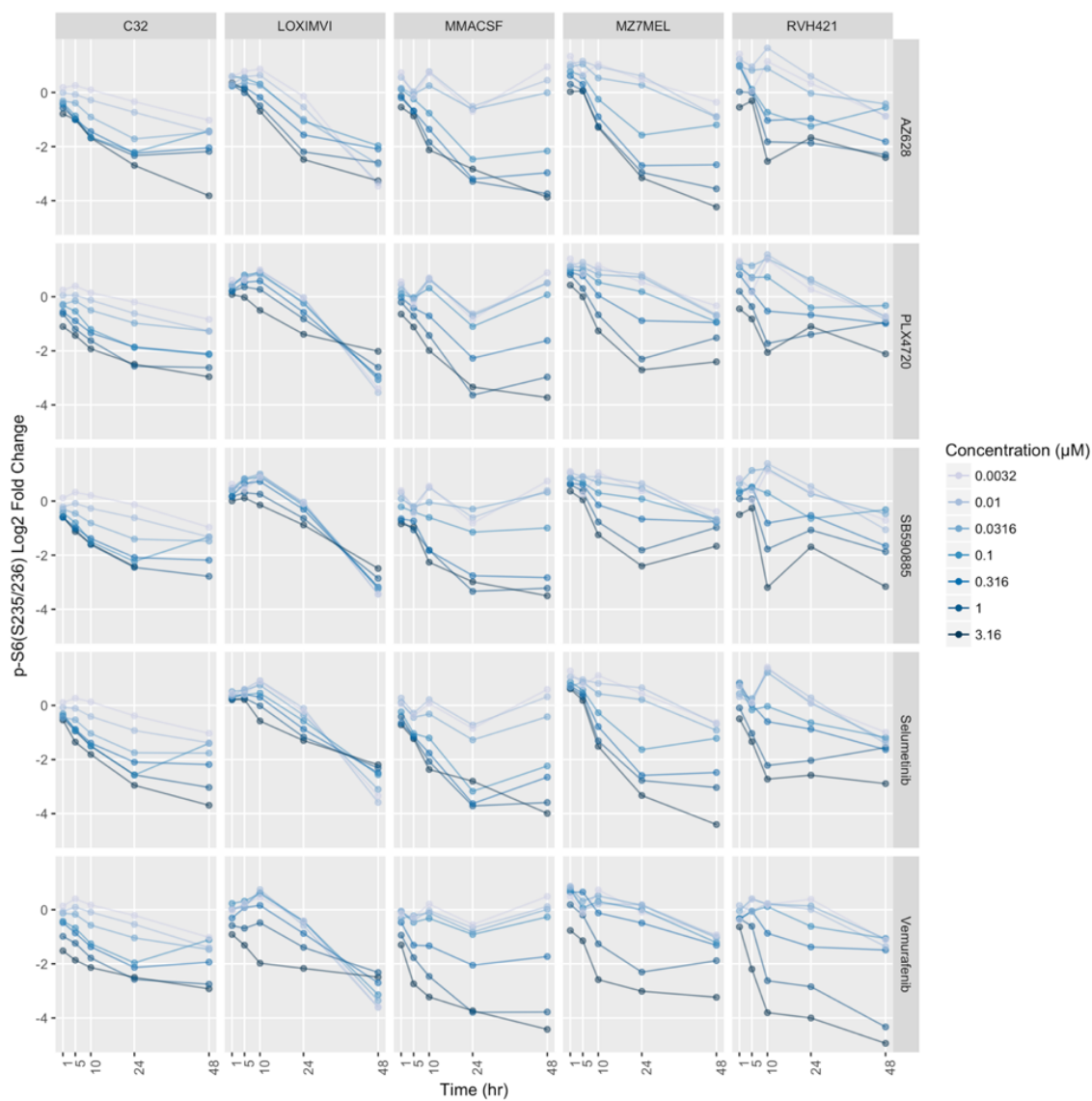

*Figure 8 Figure for Finding 1. Each row in this figure corresponds to a drug (drug labels are to the right of each row in gray boxes). Each column corresponds to a cell line (labels in gray boxes at the top of the plot). We observe a dose-dependent decrease of p-S6(S235/236) across five cell lines and drugs.*

# Appendix B: Experimental Results on Human-Machine Cooperative Curation

## Introduction

A goal of MITRE's work on evaluation has been to create a reference standard that can be used to compute how accurately systems and/or humans read and curate the literature. In the Big Mechanism Phase I evaluation, model-relevant interactions were captured from papers in three ways: by machine systems, through human-edited output of machine-generated results, and by human-curation only. For human-curation, eight curators wrote index cards for three to 17 papers each. The purpose of including human curation was to provide a baseline for comparison to machine output and to potentially create a "silver standard" – a set of index cards that are correct, but not a complete set of all possible correct index cards for the papers. If the curator-generated cards were very accurate, then they could potentially be used as a silver standard to partially evaluate recall for machine-generated cards. To assess this, we evaluated the accuracy of curator-generated cards, and compared the kinds of errors on curator-generated cards to error types on cards generated by machines. The results showed that curator-generated cards would need substantial editing to make them accurate, particularly in grounding (see below, Table B1 in Results and Discussion).

This write-up describes a follow-on experiment, which asks whether machines can improve upon human-generated index cards to facilitate the building of a silver standard. The experiment was done in two parts. For Part I, machines were given the *sentence* that a human curator selected as evidence for an index card. In Part II, we asked whether machines could edit human-generated cards to improve upon human-only results. To do this, we provided the machines with the actual index card generated by the human curator as well as the sentence human curators selected. The Part II experiment is the inverse of the experiment performed at the Phase I evaluation, which evaluated human edits of machine generated cards.

We hypothesized that if the results of this experiment were positive, then we could explore the creation of a hybrid system, combining results from multiple sources (machines and human curation), to derive a "silver standard" suitable for larger scale Big Mechanism reading experiments. Part I results showed that combining output from human generated and machine generated cards yielded correct matches to 48% of the silver standard cards. Interestingly, in Part I the machine systems collectively generated cards that matched the reference set at frequency comparable to the overlap between the human annotators curating the same paper (Slide 10 in Appendix C). The Part II results showed that machine-editing via re-grounding of entities improved the percent of cards correct for the major fields and grounding from 7.5% for humans alone to 73% for the pooled machine-edited cards. These results indicate that combining machine abilities with human curation can greatly improve upon human curation alone; however, we still need to explore whether it is possible to automatically combine outputs to create a useful silver standard, and whether this can be done at scale.

## Methods

In the Phase I evaluation, eight human curators, of varying backgrounds, generated 356 validated cards for 37 papers; most papers were curated by two curators. We manually scored a subset of these cards using the same scoring criteria that were used to evaluate machine and human-machine cards. For each of the eight human curators, we scored five cards for a total of 40 cards. The selected cards for each curator were distributed over multiple papers and were about experimental results, as opposed to background information obtained from other studies and cited in the paper. We also generated a corrected version – a "gold card", for each of the 40 scored human-generated cards.
In the Part I experiment, the machines were given the *sentences* selected by human curators as evidence text, and asked to generate index cards for these. In Part II, the machines were given the human-generated cards and asked to edit them. Edits included re-grounding of at least some of the entities, and for one system, adding some model element links. Approaches included directly editing the cards and using matching between human-generated cards and the machine cards to make changes. For scoring for both Part I and Part II, we proceeded in two steps: first we identified the best machine-generated card (for each system) that at least partially matched each gold card for the given sentence; and then we scored the matching or partially matching cards.[1]

### Matching

A machine card was considered to be at least a partial match if either:
- All three major information fields (interaction, participant A, participant B) matched the gold card, or
- Interaction and participant B matched and participant A was blank.

For evaluating matching of participants, the participant text field was used and grounding was ignored. For matching of interaction-type, we allowed some "near equivalences" as matches – specifically: *Increases/decreases phosphorylated form*, *increases/decreases_activity*, and *adds_modification* were considered matches for each other. In some cases, one of these could be wrong based on the evidence from the sentence, in which case it was marked as a match but scored as a wrong interaction type in the detailed scoring. For example, given the sentence "ligand-stimulated EphB1 induces tyrosine phosphorylation of p52Shc through SFKs and at least another tyrosine kinase", "EphB1 increases phosphorylated form of p52Shc" is correct, but "EphB1 adds modification, phosphorylation to p52Shc" is considered as a match, but marked as an interaction type error because the sentence indicates that EphB1 acts indirectly rather than directly. Similarly, a machine card that read *participant A decreases participant B* was considered a match to a gold card that said *participant A increases participant B*, but was marked as having an interaction type error. Participant A and participant B were allowed to be switched for interaction types *binds* and for *translocates.* In cases where there was more than one matching card for a gold card from a submitter, the best matching card was chosen.

---

[1]This evaluation was based on matching against a single "gold" index card associated with a specific evidence sentence from the text. Because a single sentence may describe multiple interactions, the system output often included multiple cards for the given sentence. Therefore, we performed the initial matching step to identify matches to the gold cards for scoring, and ignored the rest of the cards.

## Scoring

Error scoring was done as in the Phase I evaluation, except that cards with a missing participant A for *increases/increases_activity* were not thrown out, as they can contain useful information in the context of an ensemble. Rates for different types of errors, such as incorrectly capturing interaction types or grounding, were calculated. These are conditional scores, because the calculation was done using only the subset of matching cards that contained scoreable information for the error type. For participant and interaction type errors, only cards that were matches to the gold card and utilized the given sentence were scored. For grounding error rate and in-model error rate, all participants on matching cards that were correctly identified in the entity text field and had groundable entity types (e.g. proteins and chemicals) were scored. Error rates are not reported for *NOT True/False* and for *Relationship to Model* because there was only one NOT card in the gold standard set, and only six gold cards that were not Extension. Error rates for features such as phosphorylation and binding sites have not been calculated yet.

To investigate the potential impact of combining machine cards with human cards, we identified cases where there was a fully correct participant A (participant was correct and correctly grounded), fully correct participant B, and correct interaction type from *any* combination of the human and machine cards that matched a gold card. This represents an upper bound on the number of correct cards that could be output by an ideal ensemble system. We calculated the number of gold cards for which there were correct combinations from the human and machine cards from all submitters together, from human cards only, and from machine cards only. We also calculated the number of cards that were correct for the major fields (participant A, B, Interaction Type and Subtype, with correct grounding) generated by Part I and Part II.

# Results and Discussion

## Human Curation Results

Among the 40 scored cards generated by *human* curators, there were just three cards that correctly identified and grounded both participant A and B and had a correct interaction type. This was mostly due to poor grounding by the human curators. Entities were grounded correctly only 36% of the time. One reason is that 21% of the cards were grounded to GO instead of UniProt or HGNC (perhaps due to unclear instructions to the curators). Another 24% of groundable entities were not grounded at all. Furthermore, curators frequently used UniProt identifiers for the wrong species (e.g. gorillas, frogs, opossums, squirrels).

Curators did well at accurately capturing interactions and identifying both participants, but made errors in not including protein features in interactions. There were also errors that appear to be careless mistakes, such as including evidence text from a different paper or listing a UniProt ID for a different protein than in the entity text field. In addition, for several cards, the evidence text selected by curators did not provide all the information they annotated, suggesting that curators were using other information gained from reading the paper that they did not include in evidence. Finally, none of the curators included model element links on their cards.

**Table B1. Results for the 40 scored human-generated cards.**

| Information Type | Number of Cards | Fraction of 40 Cards Correct |
|---|---|---|
| Evidence Matched Text | 39 | 97.5 |
| Participant A Correctly Identified | 39 | 97.5 |
| Participant A Identified and Grounded Correctly | 11 | 27.5 |
| Interaction Type Correct | 35 | 87.5 |
| Participant B Correctly Identified | 38 | 95.0 |
| Participant B Identified and Grounded Correctly | 13 | 32.5 |
| Correct for All of These Fields | 3 | 7.5 |

In contrast, machines did substantially better in grounding entities in the Phase I evaluation. Among the machine-generated cards that were scored from the Phase I evaluation (note that these came from a different set of papers and were for different interactions than the human-curated cards), grounding error rates ranged from just 4% to 29%. Furthermore, machine output from three consortia included accurate model element links, which no human-curated cards had. On the other hand, machine-generated results included substantially more cards with information that was scored as largely wrong, or as having a wrong or missing participant; the top three machine systems had these errors on 31-38% of scoreable cards.

## Part I Results

Given the sentence selected by human-curators, machine systems collectively found the same interaction (same participant A, B and interaction type, not considering grounding) for 20 (50%) of the gold card interactions (Tables B2 and B3; there is a supplementary file that lists matched and unmatched gold cards with the evidence sentences). Machine systems also had partial matches – participant B and the interaction matched, but participant A was left unidentified- for an additional 8 gold cards. The best performing machine system found the same interaction (same participant A, B and interaction type, not considering grounding) as on the gold card in 32% (13/40) of the cases (Table B2). When partial matches are taken into account, the best performing machine system identified half of the gold card interactions.

**Table B2. Reference set overlap for machine-generated cards.**

| Submitter | Number of Full Matches | Number of Partial Matches | Fraction of Reference Set with a Full or Partial Match | Number of Matching Cards that were Correct for the 3 Major Fields and Grounding |
|---|---|---|---|---|
| System A | 13 | 7 | 0.50 | 7 |
| System B | 8 | 6 | 0.35 | 2 |
| System C | 3 | 6 | 0.23 | 2 |
| System D | 3 | 2 | 0.13 | 3 |

The collective recall of all the machine systems for the gold card reference set (50% for full matches) was similar to the level of overlap between human curators. Among the papers that were doubly or triply curated, there were 35 unique gold cards. Of these, 19 (54%) interactions were identified by more than one human curator. Of these 19 cards, machine systems identified full matches for 12 (68%). Collectively, the machine systems wrote cards that were correct for participant A, B, and interaction type with correct grounding for 12 of the 40 gold cards, which was higher than the 3 generated by humans alone.

It is important to note that these results only include machine cards that used the same sentence as evidence. Machine results from whole paper reading could discover the same interaction from other sentences, and potentially substantially increase the reference set overlap.

**Table B3. Matches to individual gold cards by each machine submitter.** Full matches: "2"; Partial matches: "1"; No match:"0". Participant A, B, and Interaction are from the gold card.

| Sentence ID | System A | System B | System C | System D | Participant A Text | Interaction | Participant B Text |
|---|---|---|---|---|---|---|---|
| 78 | 2 | 2 | 2 | 0 | Grb7 | binds | EphB1 |
| 147 | 2 | 2 | 1 | 0 | c-Src | binds | IQGAP1 |
| 71 | 2 | 2 | 0 | 0 | p85 | binds | Abi1 |
| 95 | 2 | 2 | 0 | 0 | EPHB1 | increases | p52^Shc |
| 4 | 2 | 1 | 1 | 0 | mTOR | adds_modification | p53 |
| 24 | 2 | 0 | 2 | 0 | Akt1 | adds_modification | Cby |
| 12 | 2 | 0 | 0 | 0 | ATR | binds | ATRIP |
| 54 | 2 | 0 | 0 | 0 | (blank) | adds_modification | Egfr |
| 64 | 2 | 0 | 0 | 0 | ZAP70 | adds_modification | pp29/30 / TRIM |
| 93 | 2 | 0 | 0 | 0 | SMYD3 | adds_modification | MAP3K2 |
| 180 | 2 | 0 | 0 | 0 | ERK1 | adds_modification | FBW7 |
| 182 | 2 | 0 | 0 | 0 | Eps8 | binds | E3b1 |
| 213 | 2 | 0 | 0 | 0 | HDAC3 | decreases | GM-CSF |
| 91 | 1 | 1 | 1 | 2 | EGFR | adds_modification | Shc |
| 14 | 1 | 1 | 0 | 0 | LAT | increases | p38 |
| 34 | 1 | 1 | 0 | 0 | BCR | translocates | Bam32 |
| 52 | 1 | 1 | 0 | 0 | ITGB1 | adds_modification | FOXO1 |
| 181 | 1 | 0 | 0 | 2 | SYK | adds_modification | HS1 |
| 188 | 1 | 0 | 0 | 1 | EGFR | adds_modification | EGFR |
| 87 | 1 | 0 | 0 | 0 | Ret9 | increases_activity | AKT1 |
| 168 | 0 | 2 | 2 | 0 | VEGFR-2 | binds | c-Src |
| 9 | 0 | 2 | 1 | 0 | EphB1 | increases_activity | ERK1 |
| 163 | 0 | 2 | 0 | 0 | arsenic | increases | cdk4 |
| 165 | 0 | 2 | 0 | 0 | Bam32 | binds | PI(3,4,5)3 |
| 55 | 0 | 1 | 0 | 1 | Fyn | increases_activity | Pyk2 |
| 83 | 0 | 0 | 1 | 0 | YWHAZ | binds | Cby |
| 143 | 0 | 0 | 1 | 0 | SMYD3 | increases_activity | MAPK1 |
| 13 | 0 | 0 | 0 | 2 | Cas | binds | FAK |

To investigate the extent that machine-generated cards could contribute to improving on a human card, we examined error rates for machine cards that at least partially matched a gold card. We also calculated error rates for the *uncorrected* human cards corresponding to the gold cards with machine matches.[2] Table B4 shows that while the machine systems had high participant error rates, which occurred because they frequently did not identify participant A, machines did substantially better at correctly grounding entities.

**Table B4. Conditional error rates for different types of errors for machine cards that at least partially matched a gold card and the corresponding set of human-curated cards.**

| | Human-generated | | System A | | System B | | System C | | System D | |
|---|---|---|---|---|---|---|---|---|---|---|
| | Error Rate | Number Scored | Error Rate | Number Scored | Error Rate | Number Scored | Error Rate | Number Scored | Error Rate | Number Scored |
| **Participant Error Rate** | 0.04 | 28 | 0.35 | 20 | 0.43 | 14 | 0.67 | 9 | 0.40 | 5 |
| **Interaction Type Error Rate** | 0.07 | 28 | 0.05 | 20 | 0.14 | 14 | 0.33 | 9 | 0.00 | 5 |
| **Grounding Error Rate** | 0.64 | 55 | 0.18 | 33 | 0.27 | 22 | 0.08 | 12 | 0.00 | 8 |
| **In Model Error Rate** | 0.11 | 55 | 0.21 | 33 | 0.23 | 22 | 0.08 | 12 | 0.00 | 8 |

It is important to note that these error rates are conditional on matching, and do not include errors severe enough to result in mismatching. Consequently, these are not equivalent to overall error rates for machine cards. The number of instances scored varies among information types and submitters depending on the number of matching cards and the number of scoreable fields.

---

[2] There were 28 gold cards with machine matches, so this was used as the pool of scoreable cards for comparison. Since each card had one or two participants, the total number of groundable entities was 55 for these 28 cards.

To measure the improvement that machine cards could potentially make when pooled with human cards, we evaluated the number of cases where human and matching Part I machine cards from all the submitters together had a correct combination of correctly identified and grounded participant A, participant B, and interaction type (see Table B5 for examples). When human and matching machine cards were combined, there were 19 gold cards for which there were there were fully correct combinations for these major fields (Table B6). Thus adding in machine cards brought the potential of getting all three major fields correct up from just 7.5% for human curators alone and 38% for combinations from machines alone, to 48% for human curation combined with the machine outputs (across all systems). For gold cards for which there was at least one partially matching machine card, there was a correct combination from machine and human output 71% of the time.

**Table B5. Example results about correct combinations for the major information fields.**

| Sentence ID | Number of Matching Machine Plus Human Cards | Number of Cards with Participant A Correct | Number of Cards with Participant B Correct | Number of Cards with Interaction Type Correct | Is there a Correct Combo? |
|---|---|---|---|---|---|
| 91 | 5 | 1 | 4 | 4 | Yes |
| 188 | 3 | 0 | 2 | 3 | No |
| 12 | 2 | 1 | 1 | 2 | Yes |

## Part II Results

Machine-editing improved the percent of cards correct for the major fields and grounding from 7.5% for humans alone to 73% (Table B6). The only machine edits were re-grounding of entities. With re-grounding, the machine systems were collectively able to generate cards that were correct for participant A, B, and interaction type with correct grounding for 29 of the 40 gold cards (Table B6.) Among the eleven gold cards that were missed, six had human errors in Participant A, B and/or interaction type that were not correctable by just re-grounding, and five cards were not grounded correctly by any of the teams. The entities on the cards that all the teams failed to ground correctly were "p38 kinase", "14-3-3", "P85", "Ca2+/calmodulin" (need to ground both), "p52^Shc", and "PI(3,4,5)3". This list includes non-specific and non-standard protein identifiers, a complex, and a chemical, some of which would require information beyond the sentence to ground correctly.

Grounding error rates varied substantially among the machine systems (Table B7). Two of the systems kept a portion of the human grounding rather than overwriting all cases. When machines overwrote human grounding, they fared better. In addition, three of the four machine systems edited cards directly, while one system used matching of machine output and only submitted cards for a subset of the gold cards.

**Table B6. Number of correct combinations and correct cards for the major information fields from human-only, Part I, and Part II.**

|  | Number of gold cards for which there was a correct card for the 3 major fields | Fraction Correct of the 40 gold cards |
|---|---|---|
| **Human curators only** | 3 | 0.08 |
| **Machines only Part I (correct cards)** | 12 | 0.30 |
| **Machines only Part I (correct combinations)** | 14 | 0.35 |
| **Human and machine cards Part I together (correct combinations)** | 19 | 0.48 |
| **Machine only Part II** | 29 | 0.73 |
| **Human-generated cards that were correct for the 3 major fields without considering grounding** | 34 | 0.85 |

**Table B7. Numbers of cards correct for the major fields and grounding error rates for submitters in Part II.**

| Submitter | Number of Matching Cards Generated | Number that were Correct for the 3 Major Fields and Grounding | Grounding Error Rate | Fraction Now Correct for the 40 Gold Cards |
|---|---|---|---|---|
| System D | 38 | 26 | 0.09 | 0.65 |
| System A | 29 | 12 | 0.35 | 0.30 |
| System B | 38 | 6 | 0.53 | 0.15 |
| System C | 4 | 2 | 0.25 | 0.05 |
| Human curators | 38 | 3 | 0.65 | 0.08 |

# Conclusions

These results indicate that combining machine abilities with human curation can greatly improve upon human curation alone. In particular, in creating a reference set, we should incorporate a high-performing automated grounder rather than having humans do it. The grounding by machines can be checked by humans for errors, using automated output of information about UniProt ids that the MITRE team has already set up. Furthermore, we now know that certain types of entities (e.g. non-specific names like p85 and complexes) will require human intervention in grounding given the best current grounding capabilities.

The results also suggest that machine systems have the ability to achieve recall levels at least comparable to novice human curators. Given that whole paper reading should enable the capture of additional interactions beyond single sentence reading, the prospects for machines exceeding the number of interactions humans can identify in a limited amount of time from a single paper are good. However, with the current technology, the machines still have substantially high error rate rates for identifying participants.

The results indicate several areas for improvement in machine reading, particularly in identifying both participants in interactions, and in using context to ground entities that may require information beyond the sentence. It would also probably be useful to identify characteristics of sentences and interactions that are correctly captured by machines versus those that are not. The later may indicate additional areas to improve upon in reading. The comparison of the gold cards with whole paper results from the Phase I evaluation suggests that whole-paper reading will improve recall for a reference set.

# Appendix C: Exploring Human-Machine Cooperative Curation

This section contains a presentation about the experiments described in Appendix B, presented at the 2016 Biocuration conference.

## Exploring Human-Machine Cooperative Curation

Tonia Korves, Christopher Garay, Robyn Kozierok,
Matt Peterson and Lynette Hirschman
The MITRE Corporation

Biocuration 2016
Geneva, April 11-14, 2016

[1]This work has been funded under DARPA contract W56KGU-15-C-0010

MITRE

| 2 |

## Problem

- **Expert curation can not keep up with the huge volume of published literature**
- **This makes it critical to find ways to speed up and partially automate curation**

- **Objective: Can humans and machines together create more accurate and efficient curation?**
  - What do people do well?
  - What do automated systems do well?
  - How are these best combined?

MITRE

# DARPA Big Mechanism Program

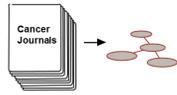
**Read** papers in cancer biology and extract causal fragments of signaling pathways, represented at all relevant semantic levels.

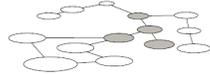
**Assemble** causal fragments into more complete pathways; discover and resolve inconsistencies.

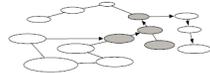
**Explain** phenomena in signaling pathways. Answer questions, including "reaching down to data," when it is available.

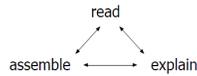
**Integrate** reading, assembly and explanation in a non-pipeline architecture that provides flexible control.

**MITRE Role:** Design and run evaluations for reading, assembly and reasoning

2014/01/31  Paul Cohen            Approved for Public Release, Distribution Unlimited

MITRE

# Big Mechanism Year 1 Evaluation Goals

- **Determine how well an automated system can read the scientific literature about a Big Mechanism**
  - Chosen domain: Ras signaling
- **Key dimensions**
  - **Capture key experimental findings from the literature**
  - **Relate these findings to an existing model**
  - **At scale (~1000 document corpus)**

MITRE

# Example Annotation

**PMC ID: PMC3902907**
**Extracted Information:**
**Participant A: JAK3**
    **Entity Type: Protein**
    **Identifier: JAK3_HUMAN**
    **Entity in Model? Yes**
**Participant B: HuR**
    **Entity Type: Protein**
    **Identifier: ELAV1_HUMAN**
    **Entity in Model? Yes**
**Interaction Type: Adds Modification**
    **Phosphorylation @ 63; 68; 200**
**Relation to Model: Corroboration**
**Text evidence:** *As identified by mass spec analysis, JAK3 phosphorylates three HuR residues (Y63, Y68, Y200)….*

Representation required:
- Representation in terms of:
  - Participant A (and features)
  - Participant B (and features)
  - Interaction type (and features)
- Grounding of participants by linkage to appropriate UniProt or PubChem ID
- Explicit statement about its relation to a defined model
- Presentation of textual evidence supporting the extracted information

MITRE

# Big Mechanism Year 1 Evaluation

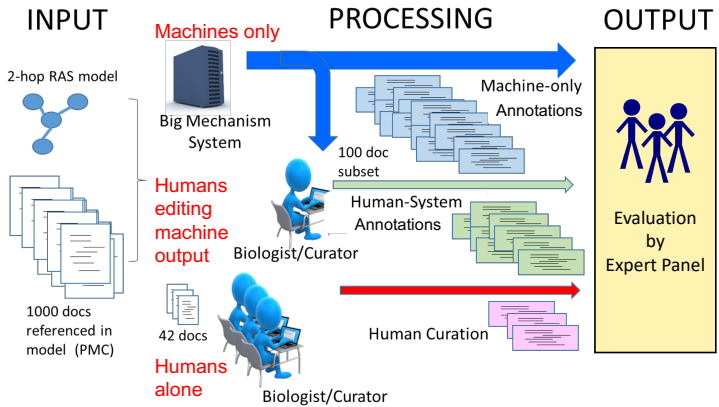

MITRE

# Human Curation

- **Motivation**
  - Determine how well machines do as compared to humans
  - Develop a reference set to evaluate automated reading
- **Experimental Methods**
  - 8 curators with varying domains of expertise, some with no prior curation experience, and very limited instruction
  - 37 papers curated by 1-3 people each
  - Selected 5 annotations per curator and scored them
- **Evaluated annotations in terms of:**
  - Correct identification of Participant A, Interaction Type, and Participant B
  - Correct grounding of entities

**MITRE**

# Human Curation Results

- **Human curators are far from perfect**
  - Lots of grounding errors
  - Some misinterpretation of sentences
  - Occasional careless mistakes in identifying participants and matching evidence text to an interaction

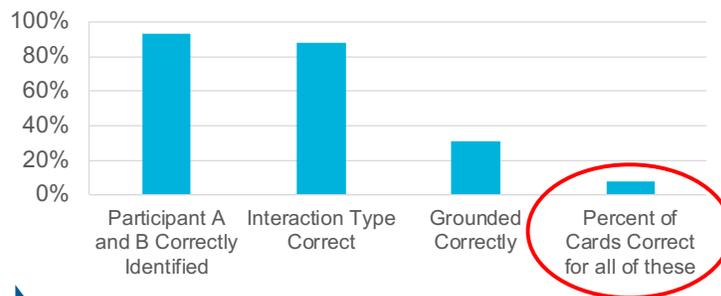

➡ **This curation, by itself, was insufficient to create a reference set**

**MITRE**

## Could Combining Human and Machine Annotations Create a Reference Set?

- **Experiment in two phases**
  - Phase 1: Machines given <u>evidence text selected by human curators</u> and generated annotations
  - Phase 2: Machines given <u>the annotations generated by the humans</u> and asked to edit these
- **Created and evaluated with a Reference Set of "Gold Cards"**
  - Gold cards created by correcting the 40 scored, human curated annotations
- **Evaluation for Phase 1:**
  - How many annotations did machines generate that "matched" gold cards (corrected versions of the human annotations)?
    - Triple Match: same Participant A and B, and a similar interaction type, not considering grounding
  - How accurate are the matching machine-generated annotations?
    - Matching annotations were scored for grounding errors, missing participant A, and interaction type errors

**MITRE**

## Phase I: Reference Set Overlap Results

- **Machine systems collectively generated triple matches (participant A, B and interaction type) for 50% of the 40 gold cards**

| Submitter | Number of Full Matches | Fraction of Gold Cards with a Full Match |
|---|---|---|
| **Machines Combined** | **20** | **0.50** |
| Machine A | 13 | 0.33 |
| Machine B | 8 | 0.20 |
| Machine C | 3 | 0.08 |
| Machine D | 3 | 0.08 |

- **The machines combined had similar odds of matching a gold card as another human curator**
  - For the 35 gold-card interactions from papers curated by more than one person, 54% were also identified by at least one other curator working on that paper

**MITRE**

## Phase I: Error Rates on Matching Annotations

- **Machines were better at grounding, but much less accurate than humans in identifying participant A than humans**

| Error Type | Human generated | Machine A | Machine B | Machine C | Machine D |
|---|---|---|---|---|---|
| Participant | 0.04 | 0.35 | 0.43 | 0.67 | 0.40 |
| Interaction Type | 0.07 | 0.05 | 0.14 | 0.33 | 0 |
| Grounding | 0.68 | 0.18 | 0.27 | 0.08 | 0 |
| Number of Annotations | 28 | 20 | 14 | 9 | 5 |

- Metrics calculated on annotations that were full or partial matches to a gold card
- Conditional error rates; sample sizes vary among groups

MITRE

## Can Machines Edit Human Annotations to Create a Reference Set?

### The Phase II Experiment

- **Machine systems were given the human-generated annotations and asked to edit them**
- **Input**
  - Human generated annotations
- **Output**
  - Machine-edited annotations
  - Edits consisted of re-grounding entities and adding some model element links

MITRE

# Phase II Results

- Machines substantially improved grounding accuracy
- Collectively, the machine editing of annotations increased the percent correct for the major fields from 8% to 73%

| Submitter | Number of Matching Annotations Generated | Number Correct for the 3 Major Fields and Grounding | Grounding Error Rate | Fraction Correct for the 40 Gold Cards |
|---|---|---|---|---|
| **Human curators** | 38 | **3** | **0.65** | **0.08** |
| **Machines Combined** | 38 | **29** | **0.07** | **0.73** |
| Machine A | 38 | 12 | 0.35 | 0.30 |
| Machine B | 38 | 6 | 0.53 | 0.15 |
| Machine C | 4 | 2 | 0.25 | 0.05 |
| Machine D | 38 | 26 | 0.09 | 0.65 |

MITRE

# Summary of Human Curation and Phase I and II Results

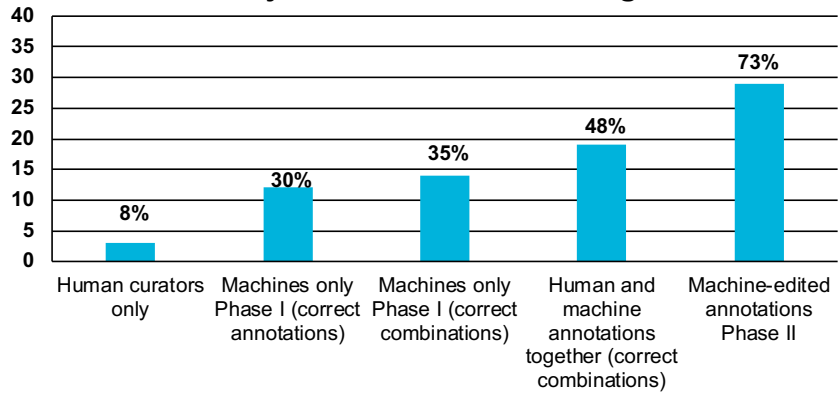

**Number of Annotations Correct for the Three Major Fields and Grounding**

- Human curators only: 8%
- Machines only Phase I (correct annotations): 30%
- Machines only Phase I (correct combinations): 35%
- Human and machine annotations together (correct combinations): 48%
- Machine-edited annotations Phase II: 73%

MITRE

# Conclusions

- **New methods for curating a reference set**
  - Use human-machine hybrid curation– Humans identify the interactions (for now), and let machines do the grounding, with some human intervention
- **Areas for improvement for machine systems**
  - Identifying participant A
  - Grounding ambiguous identifiers with whole paper reading
- **Hopeful signs of potential for machine reading**
  - Reference set overlap comparable to agreement among human curators
  - Expect to see improvements in reference set overlap and accuracy by combining annotations from whole paper reading
- **Next step: reading papers for assembly**
  - Collapsing and combining information into experimentally validated findings

**MITRE**

# Aknowledgements

- **Curators**
  - Walter Fontana, Jayasri Ganta, Jonathan Laurent, Hector Medina, Tomas Mikula, Neetu Tandon, Subha Yegnaswamy, Olga Kelmargoulis
- **Participating Teams**
  - **Reach: University of Arizona,** PI Mihai Surdeanu; Marco A. Valenzuela-Escarcega, Gus Hahn-Powell, Dane Bell, Thomas Hicks, Enrique Noriega-Atala, Clayton T. Morrison, Ryan Gutenkunst, Guang Yao, and Xia Wang
  - **NaCTEM: University of Manchester,** PIs Sophia Ananiadou and Riza Batista-Navarro; Chryssa Zerva
  - **IHMC:** PI James Allen, Lucian Galescu
  - **ISI/USC:** PI Daniel Marcu; Jose-Luis Ambite, and Sahil Garg

**MITRE**

In 1958, MITRE provided the technical know-how and operational excellence to pioneer the nation's first air defense and air traffic control systems. Today, MITRE works across government to tackle difficult problems that challenge the safety, stability, security, and well-being of our nation through its operation of federally funded R&D centers as well as public-private partnerships.

With a unique vantage point working across federal, state and local governments, as well as industry and academia, MITRE works in the public interest to discover new possibilities, create unexpected opportunities, and lead by pioneering together for public good to bring innovative ideas into existence in areas such as AI, intuitive data science, quantum information science, health informatics, policy and economic expertise, trustworthy autonomy, cyber threat sharing, and cyber resilience.

MITRE's mission-driven teams are dedicated to solving problems for a safer world.

www.mitre.org